%% file: emnlp2022main.tex
\pdfoutput=1

\documentclass[11pt]{article}

\usepackage{EMNLP2022}

\usepackage{times}
\usepackage{latexsym}

\usepackage[T1]{fontenc}
\usepackage[utf8]{inputenc}

\usepackage{microtype}

\usepackage{inconsolata}

\usepackage{booktabs}
\usepackage{graphicx}
\usepackage[section]{placeins}
\usepackage{xurl}
\usepackage{xspace}
\usepackage{longtable}

\newcommand{\shortname}{$G^3$}

\newcommand{\taskname}{Geolocation via Guidebook Grounding}
\newcommand{\numcountries}{90}

\AtBeginDocument{}%
\AtBeginDocument{}%
\newcommand\eg[0]{e.g., }
\newcommand\ie[0]{i.e., }

\title{$G^3$: Geolocation via Guidebook Grounding}

\newcommand{\aspace}{\hspace{.7em}}
\author{
        Grace Luo\thanks{\hspace{0.1cm} Denotes equal contribution.}$~^\dagger$ \aspace
        Giscard Biamby$^{*}$$^\dagger$ \aspace
        Trevor Darrell$^\dagger$ \aspace
        Daniel Fried$^\ddagger$ \aspace
        Anna Rohrbach$^\dagger$
        \\
        $^\dagger$University of California, Berkeley \quad \quad $^\ddagger$Carnegie Mellon University
        \\
        \texttt{\normalsize \{graceluo,gbiamby,trevordarrell,anna.rohrbach\}@berkeley.edu \quad dfried@cs.cmu.edu}
}

\begin{document}
\maketitle

\begin{abstract}
We demonstrate how language can improve geolocation: the task of predicting the location where an image was taken. 
Here we study explicit knowledge from human-written guidebooks that describe the salient and class-discriminative visual features humans use for geolocation. We propose the task of \emph{Geolocation via Guidebook Grounding} that uses a dataset of StreetView images from a diverse set of locations and an associated textual guidebook for GeoGuessr, a popular interactive geolocation game. Our approach predicts a country for each image by attending over the clues automatically extracted from the guidebook. Supervising attention with country-level pseudo labels achieves the best performance. Our approach substantially outperforms a state-of-the-art image-only geolocation method, with an improvement of over 5\% in Top-1 accuracy. %
Our dataset and code can be found at \url{https://github.com/g-luo/geolocation_via_guidebook_grounding}.
\end{abstract}

\input{sections/intro}
\input{sections/related}
\input{sections/dataset}
\input{sections/approach}
\input{sections/experiments}

\input{sections/analysis}
\input{sections/conclusion}
\input{sections/limitations}

\bibliography{anthology,custom}
\bibliographystyle{acl_natbib}

\input{sections/appendix}

\end{document}

%% file: sections/intro.tex
\section{\label{sec:intro} Introduction}

\begin{figure}[t!]
\begin{scriptsize}
\begin{center}
\includegraphics[width=0.95\linewidth]{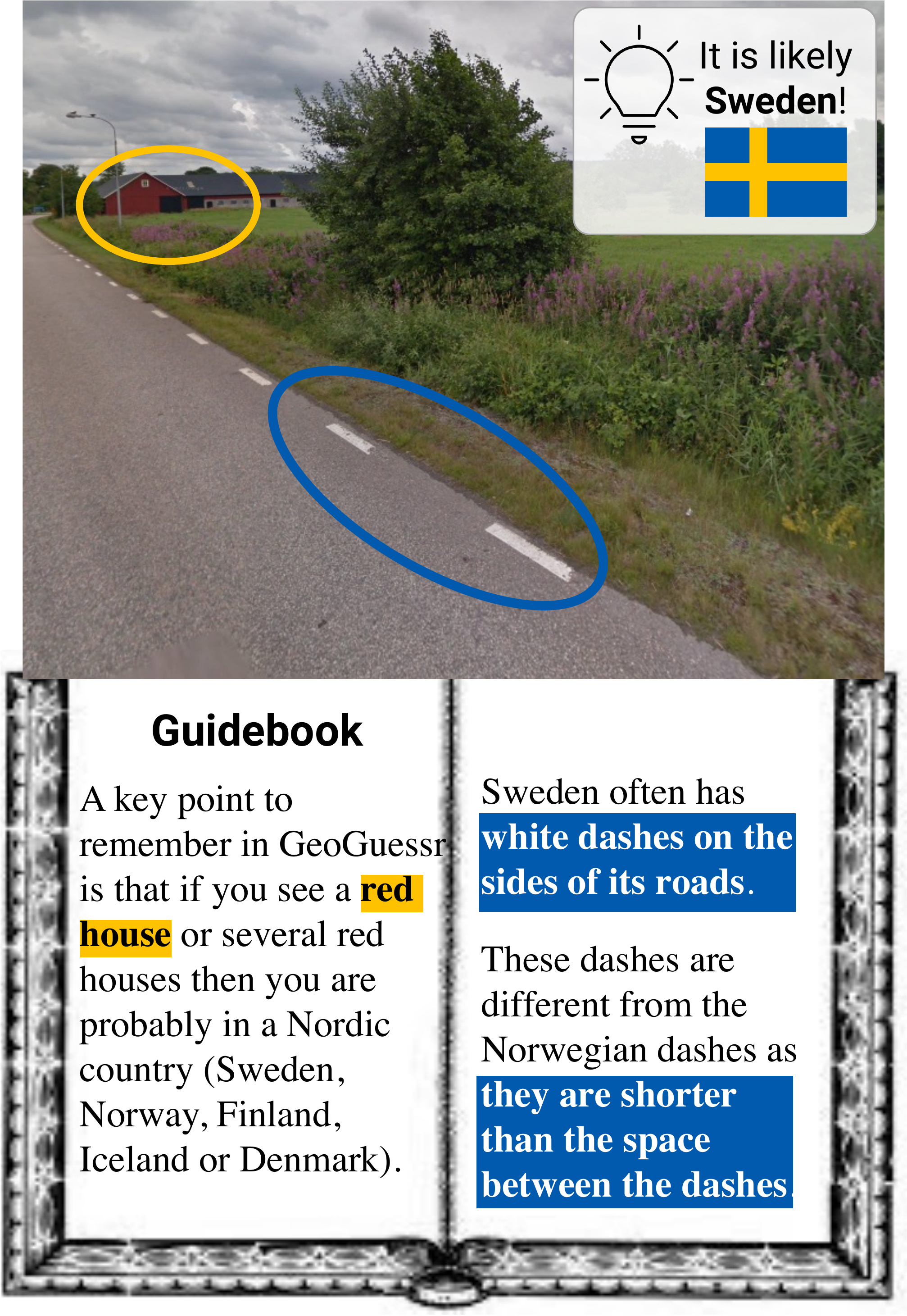}
\end{center}
\vspace{-3mm}
\caption{When asked to guess a country where an image was taken (\eg \emph{Sweden}), a person can ground the knowledge expressed in a guidebook (\eg \emph{red house}, \emph{white dashes}) to an image to inform their prediction. We propose the task of Geolocation via Guidebook Grounding, where models are tasked to do the same.}
\label{fig:teaser}
\end{scriptsize}
\end{figure}

Image geolocation plays an important role in many applications, notably in fact checking and investigative journalism, as a means of verifying or debunking claims that are illustrated with images.
For example, to verify the authenticity of video evidence of human rights abuses in Cameroon, an investigator manually matched the depicted roof coloring, building architecture, and sign text to StreetView imagery.\footnote{\url{https://www.bellingcat.com/resources/case-studies/2018/11/21/geolocation-infrastructure-destruction-cameroon-case-study-kumbo-kumfutu}} Given the rising volume of online mis- and disinformation, it is increasingly important to develop accurate automated geolocation methods. Beyond the recent importance of geolocation in fact-checking, it has also become a popular pastime in the form of online games like GeoGuessr\footnote{\url{http://www.geoguessr.com}} or Pursued.\footnote{\url{https://www.nemesys.hu/Pursued}} Since these games are so popular, human experts often publish guidebooks curating the most salient and discriminative cues for geolocating an image to teach novice players.
For example, a guide might state that \emph{Sweden often has white dashes on the sides of its roads} (Figure~\ref{fig:teaser}), knowledge a human can quickly understand and apply towards all future geolocation attempts. In contrast, most prior methods rely on training vision models on millions of images paired with GPS locations in order to learn this task \cite{kalogerakis2009image,weyand2016planet,muller2018geolocation,theiner2022interpretable}.
Recently, the authors of CLIP~\cite{radford2learning}, a large-scale multimodal model, showed that CLIP's implicit world knowledge allows it to perform geolocation once a linear classifier is added, but it does not reach the state-of-the-art performance.%

In this work we explore human-written text guides as an additional source of knowledge to complement image-based geolocation methods.
Our first contribution is the new task of \taskname{} that consists of a diverse dataset of StreetView images and an associated text guidebook created for the GeoGuessr game. The goal is to classify images into one of \numcountries{} countries, while leveraging the text clues by grounding them to the target images. Our second contribution is the proposed approach, $G^3$, which learns to leverage the guidebook. %
Namely, we combine a state-of-the-art image-only representation with our novel textual clue representation. To obtain a clue representation, we attend over the guidebook sentences (clues), while weakly supervising attention with country-specific information. Adding the final attended clue representation substantially improves performance by more than 5\% in Top-1 accuracy of a state-of-the-art geolocation model.

%% file: sections/related.tex
\section{\label{sec:related} Related Work}

\begin{figure*}[!ht]
\begin{scriptsize}
\begin{center}
\includegraphics[width=0.95\linewidth]{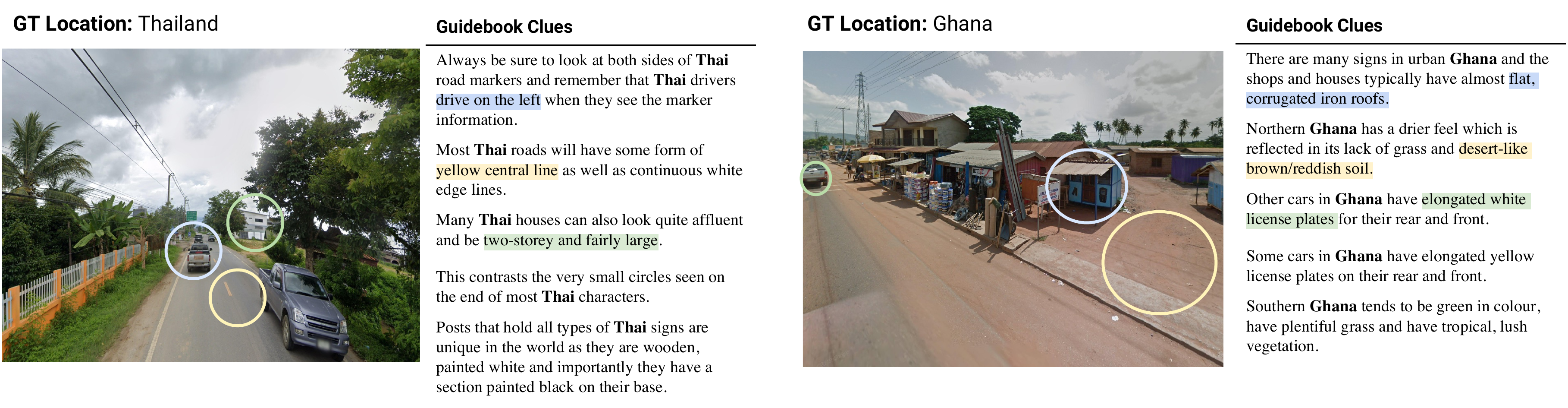}
\end{center}
\vspace{-4mm}
\caption{Examples from our dataset of StreetView Images and relevant Guidebook Text. Note how some clues are grounded in the image, for example the image in Thailand depicts cars that \textit{drive on the left}, \textit{a yellow central line}, and a house that is \textit{two-storey and fairly large}. The image in Ghana depicts houses with \textit{flat, corrugated iron roofs}, \textit{brown/reddish soil}, and a car with 
\textit{elongated white license plates}. These relevant clues were found using our country-based pseudo labels, discussed further in Section \ref{sec:approach}.}
\label{fig:pseudo_labels}
\end{scriptsize}
\end{figure*}

\textbf{Geolocation.} 
One of the first image-based geolocation methods was introduced by \citet{hays2008im2gps}, who use various handcrafted visual features to predict locations. Concretely, they 
use a nearest neighbor method against a database of labeled images. More recent works use convolutional neural networks (CNNs) rather than handcrafted features~\cite{weyand2016planet, muller2018geolocation,theiner2022interpretable}. Many works treat geolocation as a classification problem \cite{kalogerakis2009image,weyand2016planet,muller2018geolocation,theiner2022interpretable}, which requires partitioning the world map into discrete cells. 

\citet{kalogerakis2009image} discretize the map into roughly equal size rectangular bins, and \citet{weyand2016planet} use a hierarchical partitioning via a Quad tree \cite{finkel1974quad} that splits cells according to image density in order to address data imbalances. \cite{theiner2022interpretable} uses a semantic partitioning where the cells have irregular shapes that are influenced by man-made geographies (city/country borders) and natural features (rivers, mountains, etc) in order to achieve more interpretable results.
Prior works typically use large datasets of (image; GPS coordinate) pairs sourced from websites such as Flickr~\cite{hays2008im2gps,weyand2016planet,kalogerakis2009image}.

Our work differs from these in several key ways. We are the first to show the usefulness of language (\eg in the form of a guidebook) for geolocation. Moreover, in our problem statement, the task is to classify images into \numcountries{} countries. We believe this is a practical yet under-explored formulation, which allows for a more natural connection between target labels and knowledge expressed in textual form (as opposed to a somewhat arbitrary cell partitioning).
Finally, our benchmark for Geolocation via Guidebook Grounding consists of StreetView images, a more focused domain than Flickr images, which allows us to leverage the human-written guidebook for the GeoGuessr game.

\textbf{Learning with Knowledge.}
Prior works have explored using external language knowledge for various downstream tasks. For example, \citet{yang2021empirical} adopt GPT-3 to visual question answering (VQA) by leveraging the extensive knowledge of GPT-3 for answering questions given an image caption as input. \citet{marino2021krisp} use external knowledge bases and a graph neural network to boost VQA performance. Similar to a line of work in language grounding that condition on information from instructive text or game manuals to improve task performance \citep{eisenstein-etal-2009-reading,branavan-etal-2009-reinforcement,branavan2012learning,narasimhan2018grounding,andreas-etal-2018-learning,Zhong2020-rtfm}, %
we leverage a guidebook
to improve the performance of an image-only geolocation model.

\textbf{Advisable Visual Learning.}
A related line of work explores the use of language to advise vision models. \citet{kim2020advisable} show how to leverage language advice in the form of observation-action rules to better train an autonomous self-driving agent. 
In another work, a reinforcement learning soccer agent is trained from both the environment rewards and human-generated advice \cite{kuhlmann2004guiding}.
\citet{rupprecht2018guide} use inference-time language guidance to improve a trained CNN's performance on a segmentation task.
\citet{mu-etal-2020-shaping} use captions associated with images to define an auxiliary loss to improve visual features for few-shot image classification. Here, we leverage a human-written guidebook during training, however we do not have any specific instructions or advice aligned to individual images; instead we have a set of textual clues which we ground to images.

%% file: sections/dataset.tex
\section{\label{sec:dataset} \taskname{}}

\begin{figure}[tb]
\includegraphics[width=\linewidth]{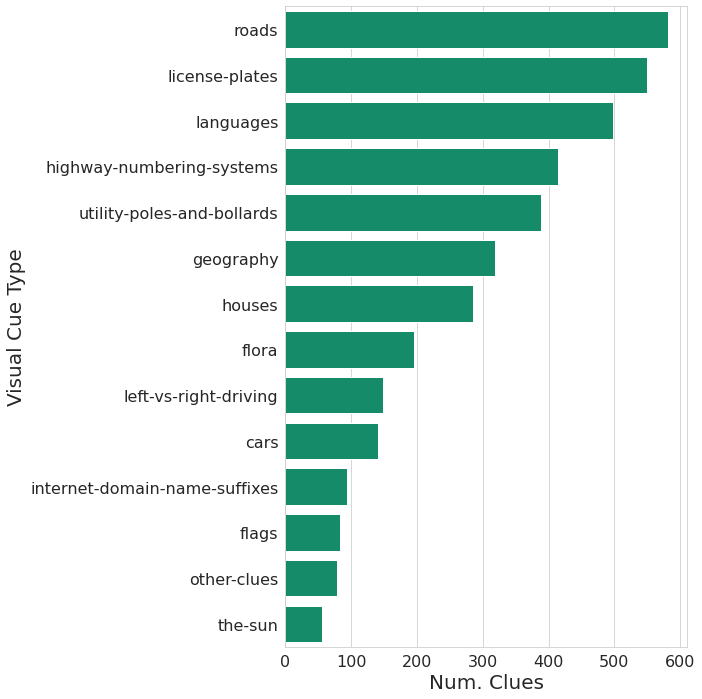}
\vspace{-5mm}
\caption{Number of guidebook clues associated with each visual cue type.}
\label{fig:clue_cluster_hist_chart}
\end{figure}

\begin{figure*}[t]
\begin{scriptsize}
\begin{center}
\includegraphics[width=0.75\linewidth]{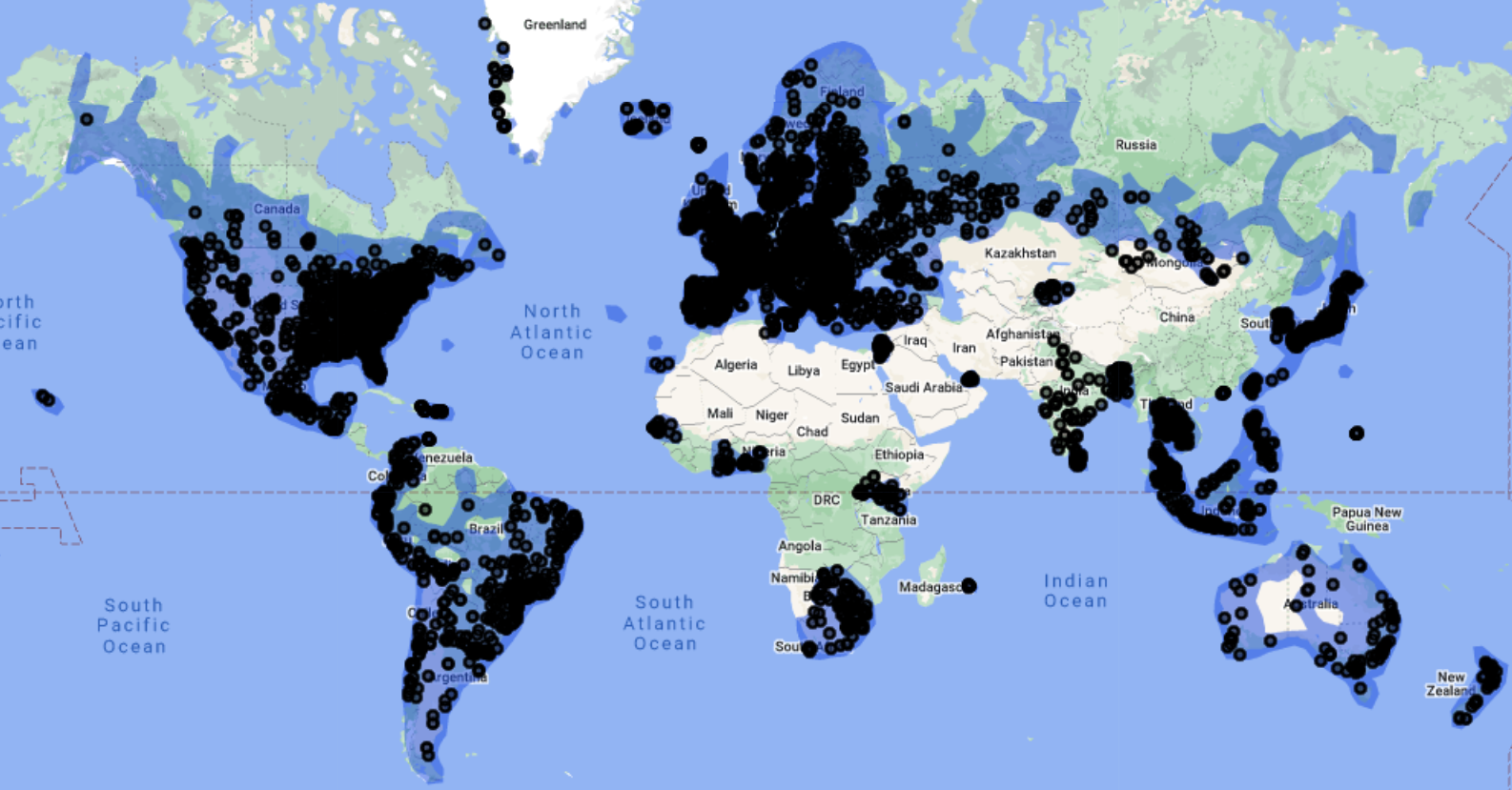}
\end{center}
\caption{Locations of 10k random panoramas in our geo-diverse StreetView dataset. Black dots denote locations present in our dataset, and blue shadings denote locations available in StreetView.}
\label{fig:streetview_coverage}
\end{scriptsize}
\end{figure*}

\begin{figure*}[t]
\begin{scriptsize}
\begin{center}
\includegraphics[width=0.9\linewidth]{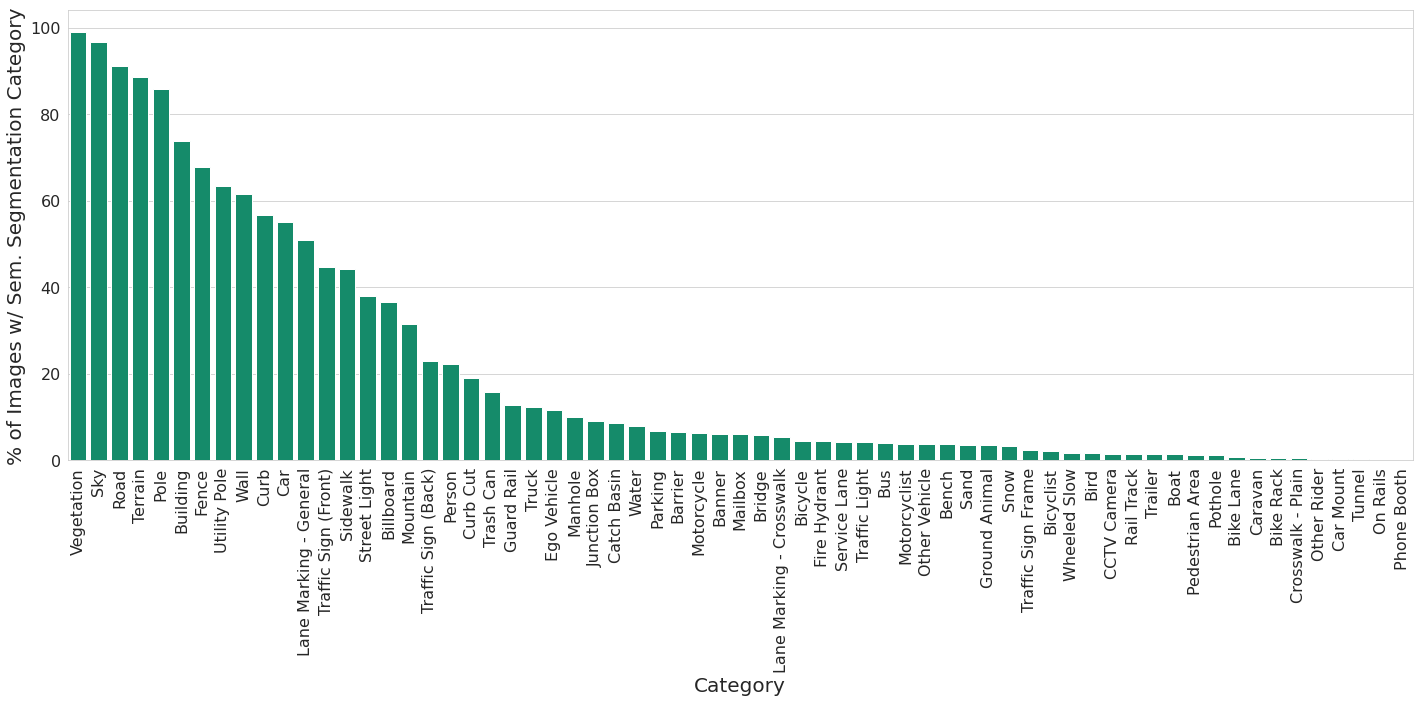}
\end{center}
\vspace{-4mm}
\caption{Breakdown of semantic segmentation categories according to what percentage of images they appear in.}
\label{fig:sem_seg_stats}
\end{scriptsize}
\end{figure*}

While prior works \cite{hays2008im2gps,weyand2016planet,larson2017benchmarking} 
propose datasets of paired images and locations, %
to the best of our knowledge, none have proposed %
datasets with images, locations, and \emph{relevant textual knowledge}. To this end, we look to guidebooks for the interactive geolocation game GeoGuessr\footnote{\url{https://www.geoguessr.com}} as a source of knowledge for the specific domain of StreetView images. For our task of \taskname{} we put together a new diverse corpus of StreetView images and an associated  text from a GeoGuessr guidebook (see examples in Figure \ref{fig:pseudo_labels}).

\paragraph{Guidebook Text: } GeoGuessr is a popular geolocation game where the user is placed into a navigable StreetView scene and must predict either the GPS coordinates or the country (depending on the gameplay mode). The game is so popular that people have built a community of online forums and guidebooks where they trade strategies for gameplay.
To take advantage of this wealth of human knowledge, we mine text from a popular guide for playing the GeoGuessr game.\footnote{\url{https://somerandomstuff1.wordpress.com/2019/02/08/geoguessr-the-top-tips-tricks-and-techniques/}} Some of the key features of this guide are that it has a broad country coverage, is well structured, and has a more ``formal'' style.
The guide is meant to teach novice players visual cues that discriminate various countries, for example the fact that \emph{Dashed white lines on the edges of roads are quite common in the countries of Denmark, Norway, Iceland and Sweden}. The guide is organized into sections for specific countries and specific visual cues, 
covering \textbf{102 countries} and over 13 visual cue types (see Figure~\ref{fig:clue_cluster_hist_chart}). 
For our final knowledge base, we select sentences from the guide that mention at least one location. We use NER tags predicted by spaCy~\cite{honnibal2017spacy} and filter for sentences with an entity tagged ``GPE'', ``LOC'', or ``NORP'', resulting in \textbf{3,832 sentences}, with an average length of 14 words.
The clue sentences contain a total of 3,712 unique words and 3,182 unique lemmas.

\begin{table}[bt]
\begin{center}
\begin{small}
\begin{tabular}{cccc}
\toprule
& Train & Validation & Test\\
\midrule
Count & 322,536 & 3,888 & 3,600 \\
\bottomrule
\end{tabular}
\end{small}
\end{center}
\caption{Our StreetView Images dataset statistics. Our validation set is roughly balanced and our test set is perfectly balanced with respect to 90 country classes.}
\vspace{-3mm}
\label{tab:dataset_stats}
\end{table}

\begin{figure*}[!ht]
\begin{scriptsize}
\begin{center}
\includegraphics[width=0.85\linewidth]{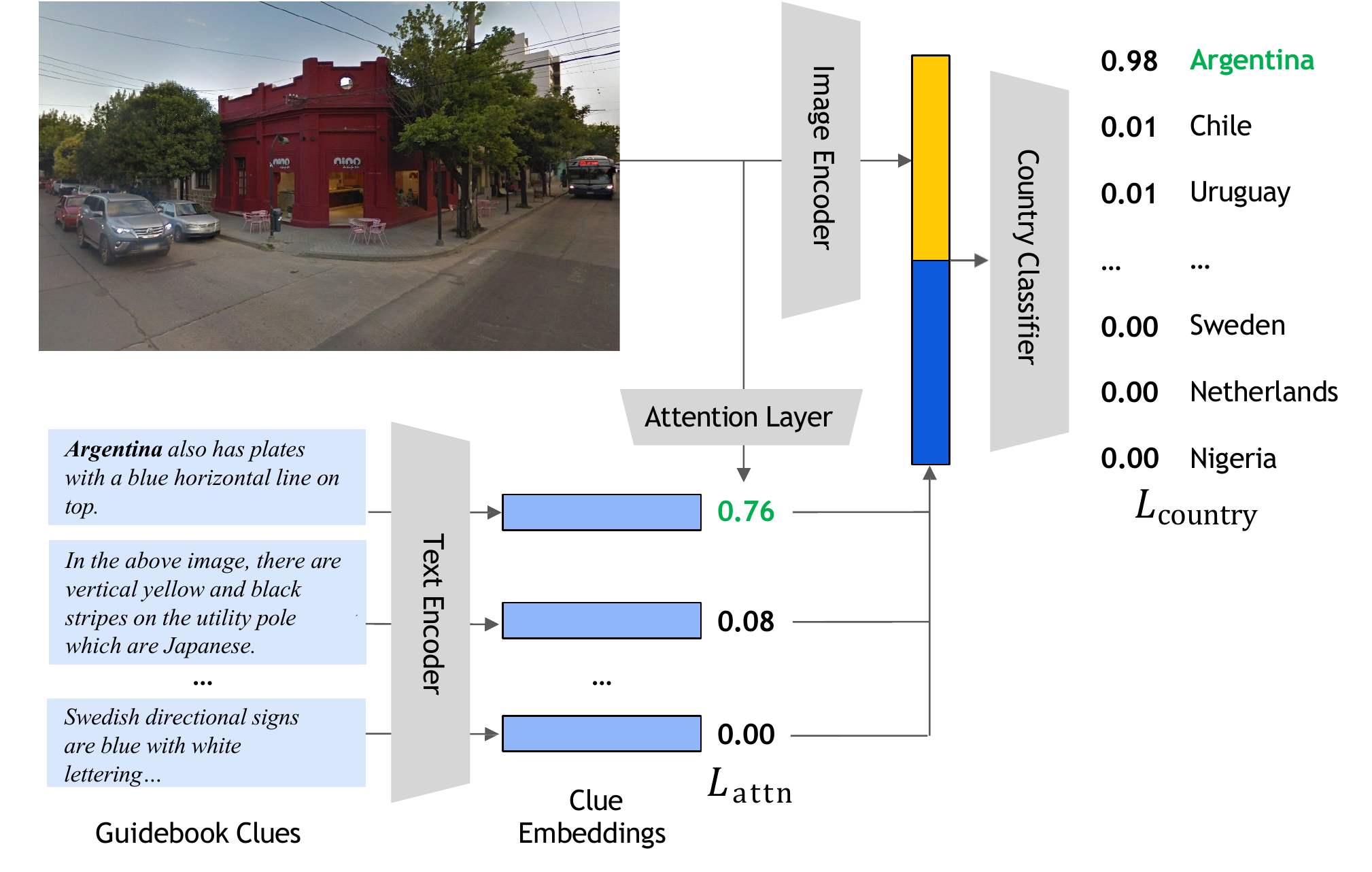}
\end{center}
\caption{Overview of our \shortname{} approach: 
We compute image embeddings for the query image, and a 
weakly supervised attention layer computes a weighted average over the clue embeddings to generate an image-relevant textual clue representation. During training the attention layer is weakly supervised with positive examples of clues that match the image's ground truth country. The image embedding is then concatenated with the clue representation before being passed to a classifier. %
}
\label{fig:approach}
\end{scriptsize}
\end{figure*}

\paragraph{StreetView Images: }
We collect a geo-diverse dataset of StreetView panoramas that covers \textbf{90 countries}. The Google StreetView API's coverage is fairly skewed, with countries in North America having some of the most extensive coverage and countries in Africa having the sparsest coverage (see Figure \ref{fig:streetview_coverage}). We gather at least \textbf{426 panoramas} per available country, as seen in Figure \ref{fig:streetview_coverage}. %
We then randomly split panoramas into training / validation / test sets. Finally, we ``cut'' each panorama into four disjoint images, as our baseline models were pretrained on images rather than 360-degree panoramas. See Table \ref{tab:dataset_stats} for the number of images in each split of the dataset. %
Our training set contains all four panorama cuts for a given image, but our validation and test sets only contain one cut to enforce independence of the evaluation samples. While our training set is imbalanced, our validation and test sets are 
balanced to ensure that each country is equally represented in the final classification performance.

We also collect semantic segmentations to better understand the content of our StreetView images (and for potential use in future work). We use MaskFormer pretrained on Mapillary Vistas \cite{cheng2021maskformer, MVD2017}, which provides segmentations for 66 categories specific to street scenes. We report the distribution of these categories present in our StreetView dataset in Figure ~\ref{fig:sem_seg_stats}. As expected, most images contain ``vegetation'', ``sky'', ``road'', many contain ``cars'' and ``lane markings'', and fewer contain ``people'', ``billboards'', ``mailboxes'',  ``fire hydrants'' etc. %

%% file: sections/approach.tex
\section{\label{sec:approach} Approach}

Our multimodal approach, \shortname, improves upon traditional image-only methods for geolocation with the help of guidebook text. We concatenate the visual representation of an image-only geolocation model with a relevant textual clue representation derived from our guidebook, and use it in a linear country classification layer, see Figure~\ref{fig:approach}.

\paragraph{Textual Clue Representation:} Given an input image to geolocate, we use a soft attention layer to compute attention scores over each sentence (clue) extracted from our guidebook to indicate its relevance. %
We opt to use the location-based attention mechanism of~\citet{luong-etal-2015-effective} as 
other attention mechanisms would take significantly more memory and time to train for a guidebook the size of ours. Our soft attention layer takes an image query $d$, encodes it with a frozen CLIP RN50x16 image encoder $f_{CLIP}$, and passes it through a fully connected layer with a ReLU activation to obtain attention logits for each clue in our guidebook: %
\begin{equation}
f_{attn}(d) = \texttt{ReLU}(W \cdot f_{CLIP}(d) + b).
\end{equation}

We precompute clue representations by applying a frozen RoBERTa Base model~\citep{roberta2019} to each clue in our guidebook.\footnote{For each sentence, we use the RoBERTa representation of the \textsc{CLS} token.} We then take the weighted average of these clue representations to obtain $\hat{G}$, an image-specific summary textual clue representation:
\begin{equation}
\hat{G} = \frac{1}{|G|}\sum_{i=1}^{|G|} \sigma{(f_{attn_{i}}(d))} \times G_{i},
\end{equation}
where we apply a sigmoid activation to the attention logits $f_{attn}(d)$ and take the $i$-th score to scale $G_i$, the $i$-th clue representation.

\paragraph{Guiding Attention with Pseudo Labels:} 
Given the large number of clues in our dataset, we use weak supervision to guide the attention mechanism during training. We create pseudo labels associating clues to images using country information. Specifically, we geoparse the clues via country demonyms and lexical matching,\footnote{We also tried the neural geoparser Mordecai (\texttt{https://github.com/openeventdata/mordecai}) but found that lexical matching had higher precision and recall.} mapping the named entities (as predicted by spaCy, see \autoref{sec:dataset}) in the clues to country labels, and therefore associated images. For example, a clue that mentions ``Japanese'' would be associated with images where Japan is the ground truth label. %
On average each image is matched to 76 country-relevant clues (2\% of all guidebook clues).
We then supervise the attention mechanism over clues with these pseudo labels. We add a binary cross entropy loss on our attention logits, where the label is a one-hot vector for each clue in our guidebook, defined by whether the clue mentions the country of the input image.
Our final loss function is as follows:
\begin{equation}
    (1 - \alpha) \times L_{country} + \alpha \times L_{attn}
\end{equation}
where $L_{country}$ is the cross entropy loss for our country classification objective and $\alpha = 0.75$ is the weighting factor for our attention loss, determined using grid search on our validation set. Since each image is only associated with a handful of clues (i.e. there are significantly more negatives than positives, or clues we do \textit{not} want to attend over), we also upweight the loss of positive pseudo labels.\footnote{We also tried supervising our attention with MIL-NCE \cite{miech2020end} but observed that our approach empirically outperformed MIL-NCE in early experiments. 
}

In summary, our pseudo labels map clues (e.g. \textit{Argentina's license plates are ...}) to countries (e.g. Argentina). At \textit{training time}, we encourage the model to attend to the clues relevant to the image's ground-truth country via the auxiliary loss $L_{attn}$. At \textit{test time}, given an input image, G3 predicts attention weights over all guidebook clues without any access to the ground-truth information.%

%% file: sections/experiments.tex
\section{\label{sec:experiments} Experiments}

\begin{table*}[th]
\begin{center}
\begin{small}
\begin{tabular}{llll}
\toprule
Model & Top-1 & Top-5 & Top-10 \\
\midrule
CLIP Nearest Neighbor & 0.4336	& 0.6858 & 0.7806\\
CLIP Linear Probe & 0.6081	$\pm$ 0.001 & 0.8789 $\pm$ 0.003 & 0.9417 $\pm$ 0.001\\
ISN &  0.6527 $\pm$ 0.015 & 0.8817 $\pm$ 0.004 & 0.9379 $\pm$ 0.004\\
$G^3$ (Ours)&  \textbf{0.7031} $\pm$ 0.002 & \textbf{0.9178} $\pm$ 0.004 & \textbf{0.9618} $\pm$ 0.002\\
\bottomrule
\end{tabular}
\end{small}
\end{center}
\caption{StreetView Image Country Classification Accuracy (Test)}
\label{tab:streetview_test}
\end{table*}

\begin{table*}[th]
\begin{center}
\begin{small}
\begin{tabular}{lllll}
\toprule
Model & Attn Supervision & Top-1 & Top-5 & Top-10 \\
\toprule
ISN &  N/A & 0.6527 $\pm$ 0.015 & 0.8817 $\pm$ 0.004 & 0.9379 $\pm$ 0.004\\
ISN + Random Text & N/A & 0.6559 $\pm$ 0.027 & 0.8840 $\pm$ 0.012 & 0.9403 $\pm$ 0.010\\
ISN + Guidebook & No & 0.6733 $\pm$ 0.011 & 0.8927 $\pm$ 0.008 & 0.9449 $\pm$ 0.005\\
ISN + Guidebook & Yes & 0.6972 $\pm$ 0.006 & 0.9115 $\pm$ 0.001 & 0.9561 $\pm$ 0.002\\
\midrule
ISN + CLIP & N/A & 0.6448 $\pm$ 0.030 & 0.8908 $\pm$ 0.011 & 0.9470 $\pm$ 0.006 \\
ISN + CLIP + Random Text & N/A & 0.6037	$\pm$ 0.035 & 0.8571 $\pm$ 0.017 & 0.9232 $\pm$ 0.010\\
ISN + CLIP + Guidebook & No & 0.6364 $\pm$ 0.037 & 0.8716 $\pm$ 0.018 & 0.9328 $\pm$ 0.013\\
\shortname = ISN + CLIP + Guidebook & Yes & \textbf{0.7031} $\pm$ 0.002 & \textbf{0.9178} $\pm$ 0.004 & \textbf{0.9618} $\pm$ 0.002\\
\bottomrule
\end{tabular}
\end{small}
\end{center}
\caption{Ablated StreetView Image Country Classification Accuracy (Test)}%
\label{tab:streetview_test_guidebook}
\end{table*}

\paragraph{Baseline Models:} For our baseline models we use ISN, a unimodal model consisting of a visual encoder and linear classification layer trained to predict a hierarchical cell on Earth given an image \cite{muller2018geolocation}, and CLIP, a multimodal model consisting of a visual and text encoder trained to maximize the cosine similarity of matched image-text pairs via contrastive learning \cite{radford2learning}. We adapt the ISN ResNet50 model pretrained on millions of Flickr images \cite{muller2018geolocation,choi2014placing} by modifying the output size of the final classification layer from the number of cells on Earth to the number of countries in our dataset and further fine-tune the model on our StreetView images. 
While CLIP was not trained specifically for the geolocation task, \citet{radford2learning} demonstrate reasonable performance on a number of geolocation benchmarks such as Countries211 \cite{radford2learning} and Im2GPS \cite{hays2008im2gps} via nearest neighbors regression, zero-shot prediction, and/or linear probing. %
We adapt the CLIP RN50x16 model pretrained on a large-scale web image-text dataset \cite{radford2learning} by taking a frozen representation from its visual encoder and %
feeding it to a linear classification layer (denoted by `CLIP Linear Probe'). We also include the nearest neighbor version (CLIP Nearest Neighbor).

\paragraph{Implementation Details:} 
In our experiments, we either use the ISN visual representation alone or concatenated with the CLIP visual embedding (``ISN + CLIP''). We do this to study the complementarity of CLIP's world knowledge to ISN.

\paragraph{Experimental Setup:}  We train our visual encoder and linear classifier at a learning rate of 1e-2 and attention layer at a learning rate of 1e-3 using an SGD optimizer (following \citealt{muller2018geolocation}), and batch size of 128 for 15 epochs. We upweight the loss for countries that appear more infrequently in our training data to account for its distributional imbalance. We also apply batch normalization on the inputs to the attention and linear classifier layers.

\paragraph{Main Results:} We report country classification accuracy on the test set of our StreetView dataset in Table~\ref{tab:streetview_test} (we report the mean and standard deviation over five seeds). Prior approaches such as ISN and CLIP demonstrate competitive performance, with 61\% and 65\% top-1 performance respectively. CLIP Linear Probe provides a significant boost of 17\% over the nearest neighbor prediction.
Finally, we observe the best performance from our full method, $G^3$, while using both the ISN and CLIP visual representations along with the domain-specific clue representations from our guidebooks, achieving a 70\% Top-1 classification accuracy.

\paragraph{Ablations:} We further ablate the effect of the different components of our approach in Table~\ref{tab:streetview_test_guidebook}.\footnote{We report the respective ablations on the validation set in the Appendix.} For both ISN and ISN + CLIP as visual representations we report the effects of using no text, attending over random text, attending over guidebook text, and including weak country supervision for the latter.
Our location-based attention mechanism introduces additional parameters into the prediction network. To disentangle the degree to which the improvements of our full method are due to its use of the guidebook text versus this additional parameterization, we perform an experiment that replaces guidebook sentences with the same amount of sentences from the news domain~\cite{biten2019good} (``Random Text'').
For the ISN model, we see that attending over random text maintains the same performance as the image-only method while guidebook text boosts the performance by 2\%. For the ISN + CLIP model, random text significantly hurts performance while guidebook text maintains similar performance within 1\%, which implies that random text can have an adverse effect and guidebook text can be redundant when combined with CLIP embeddings. 
However, when we use guidebook text and weakly-supervise the attention mechanism to encourage it to correctly select country-relevant clues, we obtain the best performance for both feature classes: 69.7\% Top-1 for ISN, and 70.3\% Top-1 for ISN+CLIP.

%% file: sections/analysis.tex
\section{\label{sec:analysis} Analysis}

\begin{figure*}[!th]
\begin{scriptsize}
\begin{center}
\includegraphics[width=\linewidth]{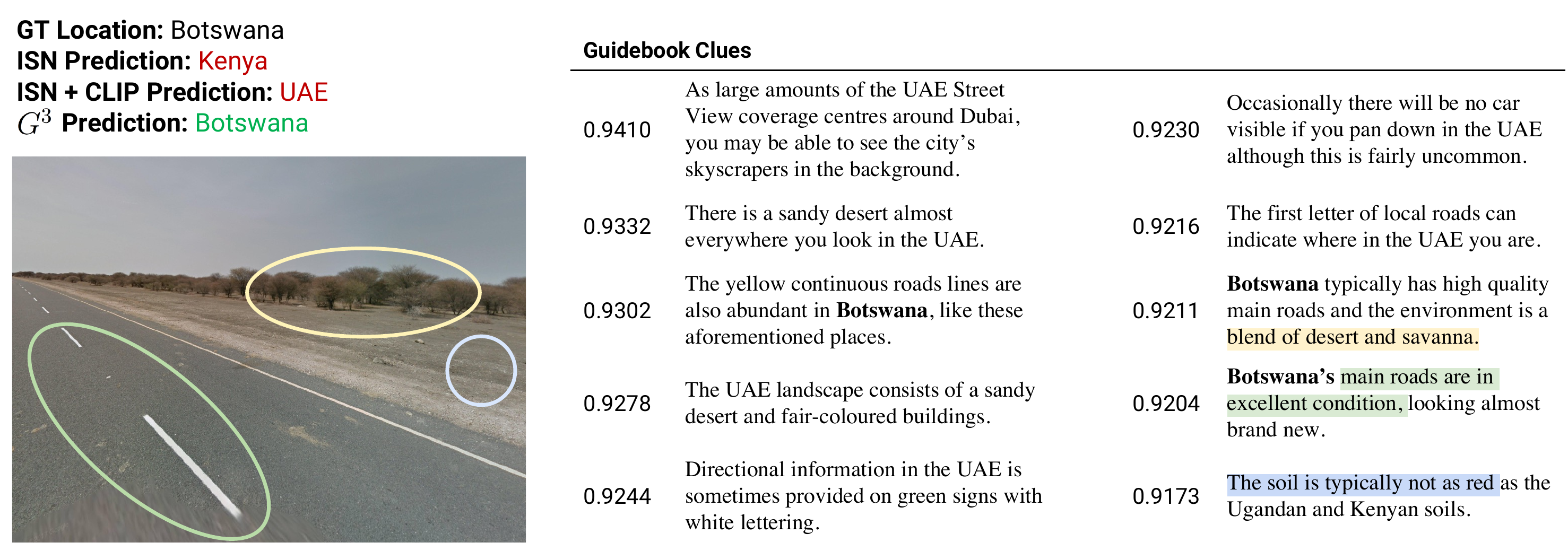}
\end{center}
\caption{An example success case from our StreetView test set. Note how ISN and ISN + CLIP make incorrect predictions, while \shortname~correctly predicts Botswana. We also depict the Top-10 clues attended over by \shortname, many of which mention relevant countries and can be grounded in the image.}
\label{fig:g3_vs_prior}
\end{scriptsize}
\end{figure*}

\begin{figure*}[!th]
\begin{scriptsize}
\begin{center}
\includegraphics[width=\linewidth]{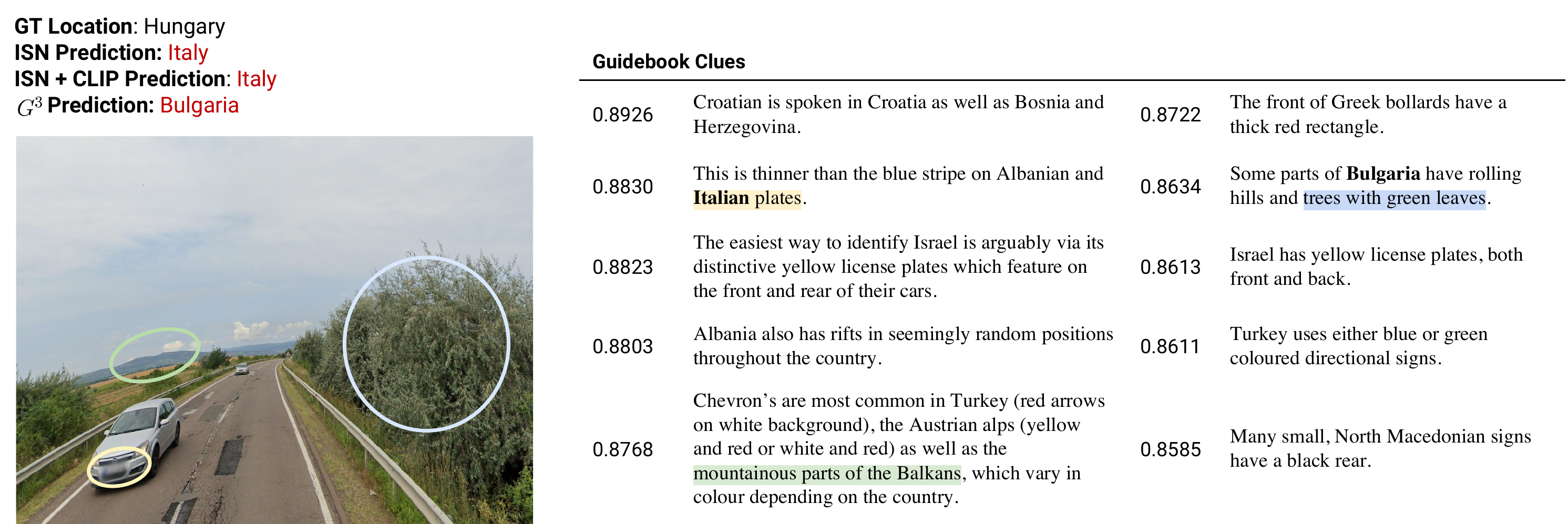}
\end{center}
\caption{An example failure case from our StreetView test set. Note how \shortname~incorrectly predicts Bulgaria, and how its Top-10 attended clues mention many countries in the Balkan Peninsula.}
\label{fig:g3_failure}
\end{scriptsize}
\end{figure*}

Here we discuss qualitative success and failure cases of \shortname~and the top clues attended over as well as comparisons with baseline image-only methods.

\paragraph{Success Cases:} In Figure \ref{fig:g3_vs_prior}, we show a qualitative example where baseline methods such as ISN and ISN + CLIP incorrectly predict Kenya and the UAE when the ground truth location is Botswana. Both incorrect predictions are plausible --- Kenya is geographically close to Botswana, and the UAE is also covered in desert. On the other hand \shortname~correctly predicts Botswana, and we visualize the Top-10 clues retrieved by our attention mechanism. \shortname~retrieves clues that mention its final prediction, Botswana, that describe how the roads are a \textit{blend of desert and savanna} and \textit{in excellent condition}. It also retrieves clues relevant to the countries it did not predict, such as the fact that the UAE often has \textit{skyscrapers in the background} and \textit{sandy desert} or Kenya often has soil that is \textit{red}. As such, our guidebook demonstrates how language can efficiently communicate what locations are commonly confused (i.e. through the co-occurence of two countries in text) and key visual cues that remedy this confusion (i.e. a country may be distinct because \textit{the soil is typically not as red} as its neighbors).

\paragraph{Failure Cases:} In Figure \ref{fig:g3_failure}, we show a failure case where all three methods --- ISN, ISN + CLIP, and \shortname, are unable to predict the ground truth location Hungary. Interestingly, while ISN and ISN + CLIP predict the same country --- Italy --- \shortname~instead predicts Bulgaria. In fact, most of our attention mechanism's top clue retrievals mention countries within the Balkan Peninsula, including Croatia, Albania, Turkey, Greece, North Macedonia, and finally Bulgaria. Many of the clues retrieved also mention objects that can be grounded in the image, including \textit{license plates}, \textit{mountainous parts}, and \textit{green leaves}. The plausibility of many of the retrieved clues also indicates how in some cases, the geolocation task is incredibly difficult with a single image and how many different countries can all be plausible given the same visual cues. Many failure cases can be attributed to this fact, and it seems to suggest 
reducing the remaining errors of existing models 
on the geolocation task requires a setting closer to that of GeoGuessr, where a scene may be navigable and grant access to multiple views of the same location. Although this setting is less realistic in the context of real-world geolocation tasks that often involve a single social media image, this potential future evaluation setting is particularly interesting because it gives rise to methods that can model uncertainty or 
enumerate the additional visual cues needed for it to be more confident in its prediction.

%% file: sections/conclusion.tex
\section{\label{sec:conclusion} Conclusion}
We presented \taskname, a new multimodal task that includes a geo-diverse dataset of images from StreetView and text from a guidebook for the GeoGuessr game.
We demonstrated that adding clues from guidebooks via our approach, \shortname, substantially outperforms past state-of-the-art image-only geolocation models on our task, with an absolute improvement of 5\% to reach 70\% in Top-1 country classification.

At the same time, there is still significant room left for improvement, and we hope that other practitioners will find this task interesting and relevant. The unique feature of our dataset is that the guidebook contains references to many nuanced and detailed characteristics of the visual scenes, posing challenging grounding problems. From recognizing specific road markings and reasoning about the scene geography, to recognizing flags, languages and other symbols, there are many skills necessary to solve the task. The focused domain of street scenes enables us to study this problem, while still being conceptually rich and capturing a multitude of diverse geographic locations. 

\paragraph{\textit{Acknowledgements.}} This work was supported in part by DoD including DARPA’s SemaFor, PTG and/or LwLL programs, as well as BAIR’s industrial alliance programs.

%% file: sections/limitations.tex
\section{\label{sec:limitations} Limitations, Ethics, and Broader Impacts}

\paragraph{Dataset: } The copyright and usage rights of the StreetView dataset are subject to that of Google. Unlike prior works that train and evaluate on datasets of user uploaded data, which are often skewed towards Western and industrialized countries, we take measures to collect a minimum number of samples from a diverse set of countries and reward performance on each country equally during evaluation. That being said, our findings are limited to the \numcountries~countries in our dataset since there exist countries that do not have Google StreetView coverage due to legal or resource reasons. However, we hope that our work contributes to a broader discussion of collecting geo-diverse data, which is important for building systems that work equally well for diverse sets of populations.

While our collection of clues used to encode human-written knowledge discusses an extensive set of countries, we note that the clues are from a guidebook with a single author and are therefore limited to the experiences of one person. We encourage future work to also consider community forums such as \url{https://www.reddit.com/r/geoguessr/} which includes discussion from a broader range of players but contains noisier data.
\paragraph{Approach: } Since we build upon pretrained models such as ISN \cite{muller2018geolocation} and CLIP \cite{radford2learning} for our visual representations, our approach is subject to the pre-existing biases learned by these models. For example, CLIP has demonstrated biases w.r.t. race and gender when classifying images of individuals \cite{radford2learning}. Our final model \shortname~is 27.9M total trainable parameters and its training takes on average 5 hours on one NVIDIA GeForce Titan X GPU, which is estimated to be 0.54 kgCO2eq in total emissions \cite{lacoste2019quantifying}.

\paragraph{Broader Impacts: } We acknowledge that approaches for the domain of geolocation can be misused in applications such as surveillance. However, we would also like to highlight that our work specifically focuses on the more coarse-level task of country classification to better align our modeling with our problem statement of language and grounding. As such, our work is less useful for fine-grained surveillance, which usually searches for locations on a city or street level. We caution against these unintended use cases, and we also emphasize that geolocation has many other positive applications in disaster response (e.g. interpreting social media imagery and appropriately directing disaster resources), arts and culture (e.g. understanding key identifying features of different locations around the world to produce more inclusive animations and films), and fact checking (e.g. determining the provenance of an image in journalism and content moderation).

%% file: sections/appendix.tex
\appendix
\section{\label{sec:appendix} Appendix}

In Section \ref{sec:appendix_val_results} we discuss results on our StreetView validation set. In Section \ref{sec:streetview_prediction_samples} we show example predictions along with the attended clues. In Section \ref{sec:appendix_guidebook_cues} we provide details about the guidebook clues.

\subsection{Validation Set Results}\label{sec:appendix_val_results}
In Tables \ref{tab:streetview_val} and \ref{tab:streetview_ablated_val} we present experimental results on our Validation set corresponding to the Test set results from Tables \ref{tab:streetview_test} and \ref{tab:streetview_test_guidebook} of Section \ref{sec:experiments}. The trends observed here are similar to those discussed previously on our Test set.

\subsection{Example Predictions with Attended Guidebook Clues}\label{sec:streetview_prediction_samples}
\begin{figure*}[h]
\begin{scriptsize}
\begin{center}
\includegraphics[width=15cm]{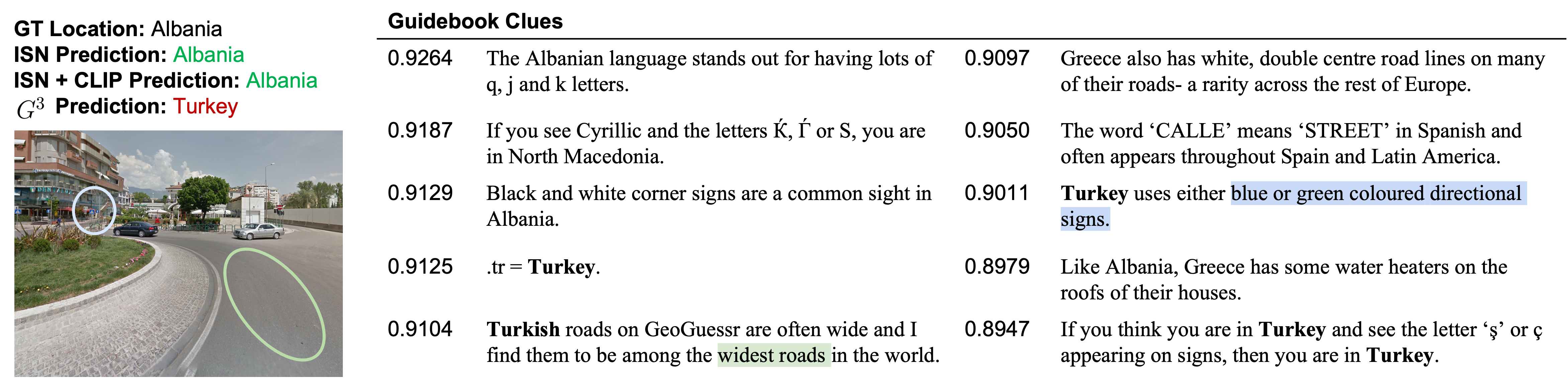}
\end{center}
\caption{Sample from our StreetView test dataset where ISN, ISN + CLIP predict correctly but \shortname~predicts incorrectly. Note how \shortname~incorrectly predicts Turkey, and how its Top-10 attended clues mention either Turkey or the ground truth country Albania.}
\label{fig:g3_failure_isn_isn-clip_success}
\end{scriptsize}
\end{figure*}

Figure \ref{fig:g3_failure_isn_isn-clip_success} shows an example where ISN and ISN + CLIP both make the correct prediction but our method \shortname~makes an incorrect prediction, which occurs 2\% of the time in our test set (whereas the inverse, where only \shortname~predicts correctly, occurs 4\% of the time). \shortname~attends to clues mentioning both the ground truth country (Albania) and its incorrect prediction (Turkey). In fact, some clues that mention Turkey can be grounded in the image, for example the \textit{[wide] roads} and \textit{blue or green coloured directional signs}. In many such failure cases, the ground truth country is mentioned in the retrieved clues but lacks visual cues that can be related to the given image.

\begin{figure*}[h]
\begin{scriptsize}
\begin{center}
\includegraphics[width=\linewidth]{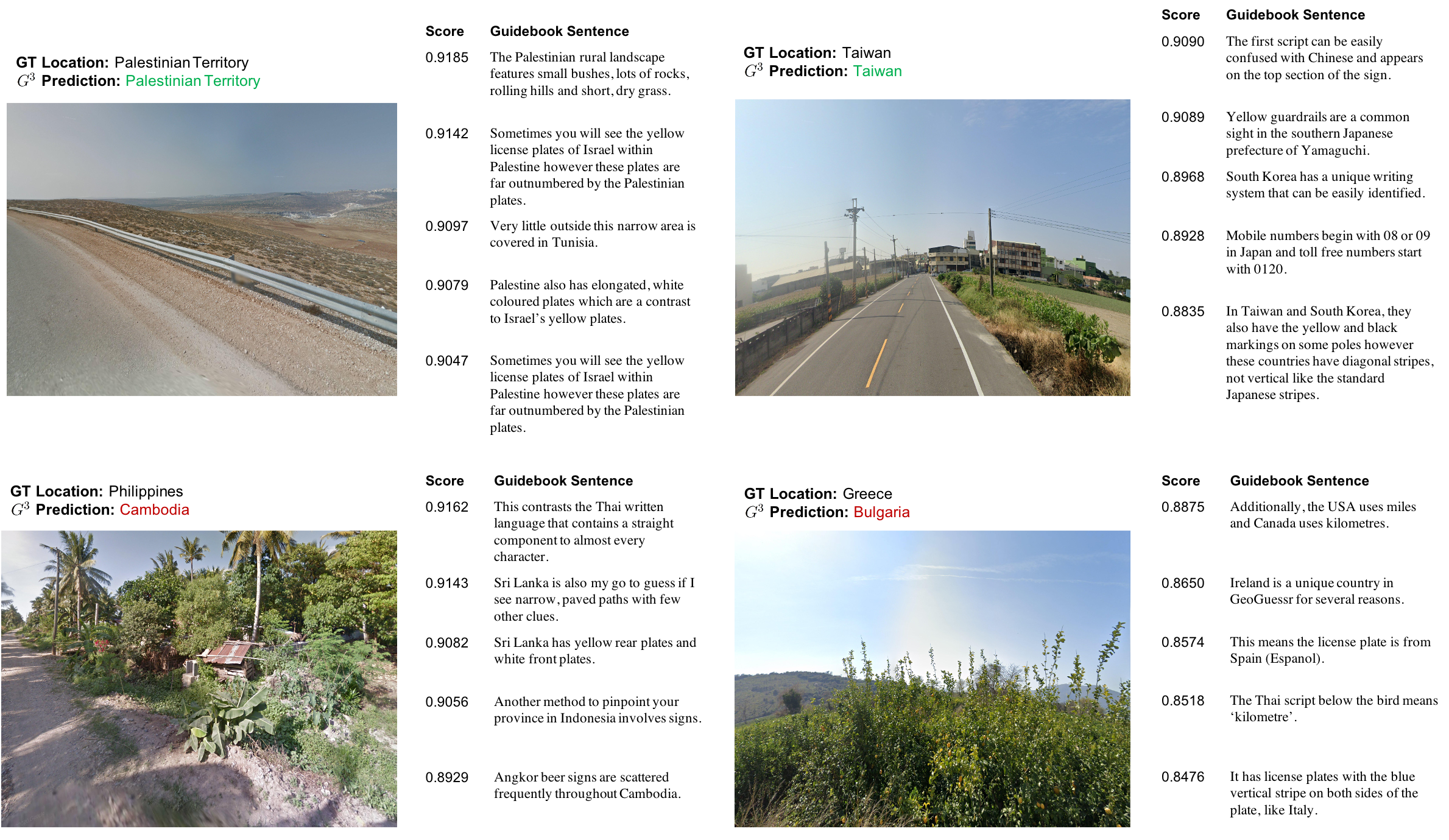}
\end{center}
\caption{Random samples from our StreetView test dataset and the Top-5 guidebook sentences attended to by \shortname.}
\label{fig:random_attn}
\end{scriptsize}
\end{figure*}

Figure \ref{fig:random_attn} shows examples of \textit{random} images drawn from our StreetView test set, the predictions of our model \shortname, and the Top-5 most relevant guidebook clues according to the model's attention scores.
In the top left example the model correctly predicts Palestinian Territory and four of the five top clues are related to Palestine. In the top right example the prediction is again correct (Taiwan), and while none of the clues mention Taiwan, they are all related to nearby geographic regions in Asia. In contrast, the bottom left example shows an incorrect prediction, though the clues relate to an overall correct geographic region. 
In the bottom right the photo shows few visual cues (\ie only vegetation is depicted) and thus the clues mention a wide spread of countries.

\subsection{Guidebook Clues}\label{sec:appendix_guidebook_cues}

\begin{figure*}[h]
\begin{scriptsize}
\begin{center}
\includegraphics[width=\linewidth]{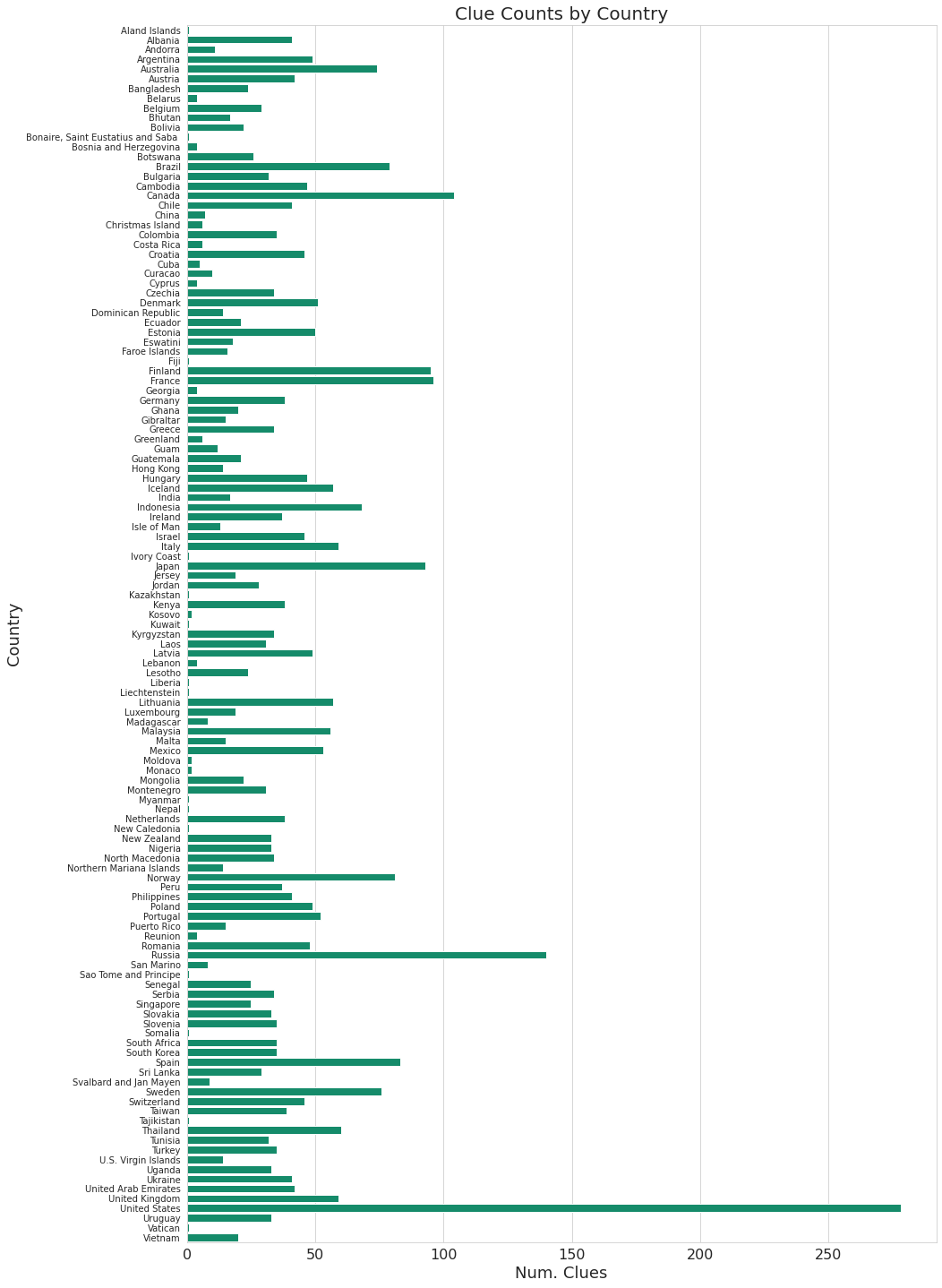}
\end{center}
\vspace{-3mm}
\caption{A histogram of the number of clues for each country.}
\label{fig:country_clue_hist}
\end{scriptsize}
\end{figure*}

Figure \ref{fig:country_clue_hist} shows a breakdown of clue counts by country. 
Some clues are not matched to a country, as in the example \textit{Birch trees are only found north of the 40th parallel,} because the clue applies to a broader geographical area rather than a country.

\begin{table*}[h]
\begin{center}
\begin{scriptsize}
\begin{tabular}{llll}
\toprule
Model & Top-1 & Top-5 & Top-10 \\
\toprule
CLIP Nearest Neighbor & 0.4313 & 0.6826 & 0.7832\\
CLIP Linear Probe &  0.6238	$\pm$ 0.001 & 0.8860 $\pm$ 0.003 & 0.9473 $\pm$ 0.002\\
ISN & 0.6548 $\pm$ 0.013 & 0.8852 $\pm$ 0.006 &	0.9442 $\pm$ 0.005\\
\shortname~(Ours) & \textbf{0.7020} $\pm$ 0.005 & \textbf{0.9224} $\pm$ 0.003 & \textbf{0.9661} $\pm$ 0.002 \\
\bottomrule
\end{tabular}
\end{scriptsize}
\caption{StreetView Image Country Classification Accuracy (Val)}
\label{tab:streetview_val}
\end{center}
\end{table*}

\begin{table*}[h]
\begin{center}
\begin{scriptsize}
\begin{tabular}{lllll}
\toprule
Model & Attn Supervision & Top-1 & Top-5 & Top-10 \\
\toprule
ISN & N/A & 0.6548 $\pm$ 0.013 & 0.8852 $\pm$ 0.006 &	0.9442 $\pm$ 0.005\\
ISN + Random Text & N/A &  0.6615 $\pm$ 0.024 &	0.8932 $\pm$ 0.012 & 0.9471 $\pm$ 0.008\\
ISN + Guidebook & No & 0.6776 $\pm$ 0.013 & 0.9007 $\pm$ 0.007 & 0.9521 $\pm$ 0.005\\
ISN + Guidebook & Yes & 0.6966 $\pm$ 0.003 & 0.9193 $\pm$ 0.003 & 0.9634 $\pm$ 0.003\\
\midrule
ISN + CLIP & N/A & 0.6576 $\pm$ 0.027 & 0.9000 $\pm$ 0.010 & 0.9549 $\pm$ 0.004\\
ISN + CLIP + Random Text & N/A & 0.6131	$\pm$ 0.029 & 0.8633	$\pm$ 0.012 & 0.9315 $\pm$ 0.010\\
ISN + CLIP + Guidebook & No & 0.6434 $\pm$ 0.036 & 0.8800 $\pm$ 0.019 & 0.9392 $\pm$ 0.013\\
\shortname = ISN + CLIP + Guidebook & Yes & \textbf{0.7020} $\pm$ 0.005 & \textbf{0.9224} $\pm$ 0.003 & \textbf{0.9661} $\pm$ 0.002 \\
\bottomrule
\end{tabular}
\caption{Ablated StreetView Image Country Classification Accuracy (Val)}%
\label{tab:streetview_ablated_val}
\end{scriptsize}
\end{center}
\end{table*}

%% file: emnlp2022main.bbl
\begin{thebibliography}{29}
\expandafter\ifx\csname natexlab\endcsname\relax\def\natexlab#1{#1}\fi

\bibitem[{Andreas et~al.(2018)Andreas, Klein, and
  Levine}]{andreas-etal-2018-learning}
Jacob Andreas, Dan Klein, and Sergey Levine. 2018.
\newblock \href {https://doi.org/10.18653/v1/N18-1197} {Learning with latent
  language}.
\newblock In \emph{Proceedings of the 2018 Conference of the North {A}merican
  Chapter of the Association for Computational Linguistics: Human Language
  Technologies, Volume 1 (Long Papers)}, pages 2166--2179, New Orleans,
  Louisiana. Association for Computational Linguistics.

\bibitem[{Biten et~al.(2019)Biten, Gomez, Rusinol, and
  Karatzas}]{biten2019good}
Ali~Furkan Biten, Lluis Gomez, Mar{\c{c}}al Rusinol, and Dimosthenis Karatzas.
  2019.
\newblock Good news, everyone! context driven entity-aware captioning for news
  images.
\newblock In \emph{Proceedings of the IEEE/CVF Conference on Computer Vision
  and Pattern Recognition}, pages 12466--12475.

\bibitem[{Branavan et~al.(2009)Branavan, Chen, Zettlemoyer, and
  Barzilay}]{branavan-etal-2009-reinforcement}
S.R.K. Branavan, Harr Chen, Luke Zettlemoyer, and Regina Barzilay. 2009.
\newblock \href {https://aclanthology.org/P09-1010} {Reinforcement learning for
  mapping instructions to actions}.
\newblock In \emph{Proceedings of the Joint Conference of the 47th Annual
  Meeting of the {ACL} and the 4th International Joint Conference on Natural
  Language Processing of the {AFNLP}}, pages 82--90, Suntec, Singapore.
  Association for Computational Linguistics.

\bibitem[{Branavan et~al.(2012)Branavan, Silver, and
  Barzilay}]{branavan2012learning}
SRK Branavan, David Silver, and Regina Barzilay. 2012.
\newblock Learning to win by reading manuals in a monte-carlo framework.
\newblock \emph{Journal of Artificial Intelligence Research}, 43:661--704.

\bibitem[{Cheng et~al.(2021)Cheng, Schwing, and Kirillov}]{cheng2021maskformer}
Bowen Cheng, Alexander~G. Schwing, and Alexander Kirillov. 2021.
\newblock Per-pixel classification is not all you need for semantic
  segmentation.
\newblock In \emph{NeurIPS}.

\bibitem[{Choi et~al.(2014)Choi, Thomee, Friedland, Cao, Ni, Borth, Elizalde,
  Gottlieb, Carrano, Pearce et~al.}]{choi2014placing}
Jaeyoung Choi, Bart Thomee, Gerald Friedland, Liangliang Cao, Karl Ni, Damian
  Borth, Benjamin Elizalde, Luke Gottlieb, Carmen Carrano, Roger Pearce, et~al.
  2014.
\newblock The placing task: A large-scale geo-estimation challenge for
  social-media videos and images.
\newblock In \emph{Proceedings of the 3rd acm multimedia workshop on geotagging
  and its applications in multimedia}, pages 27--31.

\bibitem[{Eisenstein et~al.(2009)Eisenstein, Clarke, Goldwasser, and
  Roth}]{eisenstein-etal-2009-reading}
Jacob Eisenstein, James Clarke, Dan Goldwasser, and Dan Roth. 2009.
\newblock \href {https://aclanthology.org/D09-1100} {Reading to learn:
  Constructing features from semantic abstracts}.
\newblock In \emph{Proceedings of the 2009 Conference on Empirical Methods in
  Natural Language Processing}, pages 958--967, Singapore. Association for
  Computational Linguistics.

\bibitem[{Finkel and Bentley(1974)}]{finkel1974quad}
Raphael~A Finkel and Jon~Louis Bentley. 1974.
\newblock Quad trees a data structure for retrieval on composite keys.
\newblock \emph{Acta informatica}, 4(1):1--9.

\bibitem[{Hays and Efros(2008)}]{hays2008im2gps}
James Hays and Alexei~A Efros. 2008.
\newblock Im2gps: estimating geographic information from a single image.
\newblock In \emph{2008 ieee conference on computer vision and pattern
  recognition}, pages 1--8. IEEE.

\bibitem[{Honnibal and Montani(2017)}]{honnibal2017spacy}
Matthew Honnibal and Ines Montani. 2017.
\newblock spacy 2: Natural language understanding with bloom embeddings,
  convolutional neural networks and incremental parsing.
\newblock \emph{To appear}, 7(1):411--420.

\bibitem[{Kalogerakis et~al.(2009)Kalogerakis, Vesselova, Hays, Efros, and
  Hertzmann}]{kalogerakis2009image}
Evangelos Kalogerakis, Olga Vesselova, James Hays, Alexei~A Efros, and Aaron
  Hertzmann. 2009.
\newblock Image sequence geolocation with human travel priors.
\newblock In \emph{2009 IEEE 12th international conference on computer vision},
  pages 253--260. IEEE.

\bibitem[{Kim et~al.(2020)Kim, Moon, Rohrbach, Darrell, and
  Canny}]{kim2020advisable}
Jinkyu Kim, Suhong Moon, Anna Rohrbach, Trevor Darrell, and John Canny. 2020.
\newblock Advisable learning for self-driving vehicles by internalizing
  observation-to-action rules.
\newblock In \emph{Proceedings of the IEEE/CVF Conference on Computer Vision
  and Pattern Recognition}, pages 9661--9670.

\bibitem[{Kuhlmann et~al.(2004)Kuhlmann, Stone, Mooney, and
  Shavlik}]{kuhlmann2004guiding}
Gregory Kuhlmann, Peter Stone, Raymond Mooney, and Jude Shavlik. 2004.
\newblock Guiding a reinforcement learner with natural language advice: Initial
  results in robocup soccer.
\newblock In \emph{The AAAI-2004 workshop on supervisory control of learning
  and adaptive systems}. San Jose, CA.

\bibitem[{Lacoste et~al.(2019)Lacoste, Luccioni, Schmidt, and
  Dandres}]{lacoste2019quantifying}
Alexandre Lacoste, Alexandra Luccioni, Victor Schmidt, and Thomas Dandres.
  2019.
\newblock Quantifying the carbon emissions of machine learning.
\newblock \emph{arXiv preprint arXiv:1910.09700}.

\bibitem[{Larson et~al.(2017)Larson, Soleymani, Gravier, Ionescu, and
  Jones}]{larson2017benchmarking}
Martha Larson, Mohammad Soleymani, Guillaume Gravier, Bogdan Ionescu, and
  Gareth~JF Jones. 2017.
\newblock The benchmarking initiative for multimedia evaluation: Mediaeval
  2016.
\newblock \emph{IEEE MultiMedia}, 24(1):93--96.

\bibitem[{Liu et~al.(2019)Liu, Ott, Goyal, Du, Joshi, Chen, Levy, Lewis,
  Zettlemoyer, and Stoyanov}]{roberta2019}
Yinhan Liu, Myle Ott, Naman Goyal, Jingfei Du, Mandar Joshi, Danqi Chen, Omer
  Levy, Mike Lewis, Luke Zettlemoyer, and Veselin Stoyanov. 2019.
\newblock \href {https://doi.org/10.48550/ARXIV.1907.11692} {Roberta: A
  robustly optimized bert pretraining approach}.

\bibitem[{Luong et~al.(2015)Luong, Pham, and
  Manning}]{luong-etal-2015-effective}
Thang Luong, Hieu Pham, and Christopher~D. Manning. 2015.
\newblock \href {https://doi.org/10.18653/v1/D15-1166} {Effective approaches to
  attention-based neural machine translation}.
\newblock In \emph{Proceedings of the 2015 Conference on Empirical Methods in
  Natural Language Processing}, pages 1412--1421, Lisbon, Portugal. Association
  for Computational Linguistics.

\bibitem[{Marino et~al.(2021)Marino, Chen, Parikh, Gupta, and
  Rohrbach}]{marino2021krisp}
Kenneth Marino, Xinlei Chen, Devi Parikh, Abhinav Gupta, and Marcus Rohrbach.
  2021.
\newblock Krisp: Integrating implicit and symbolic knowledge for open-domain
  knowledge-based vqa.
\newblock In \emph{Proceedings of the IEEE/CVF Conference on Computer Vision
  and Pattern Recognition}, pages 14111--14121.

\bibitem[{Miech et~al.(2020)Miech, Alayrac, Smaira, Laptev, Sivic, and
  Zisserman}]{miech2020end}
Antoine Miech, Jean-Baptiste Alayrac, Lucas Smaira, Ivan Laptev, Josef Sivic,
  and Andrew Zisserman. 2020.
\newblock End-to-end learning of visual representations from uncurated
  instructional videos.
\newblock In \emph{Proceedings of the IEEE/CVF Conference on Computer Vision
  and Pattern Recognition}, pages 9879--9889.

\bibitem[{Mu et~al.(2020)Mu, Liang, and Goodman}]{mu-etal-2020-shaping}
Jesse Mu, Percy Liang, and Noah Goodman. 2020.
\newblock \href {https://doi.org/10.18653/v1/2020.acl-main.436} {Shaping visual
  representations with language for few-shot classification}.
\newblock In \emph{Proceedings of the 58th Annual Meeting of the Association
  for Computational Linguistics}, pages 4823--4830, Online. Association for
  Computational Linguistics.

\bibitem[{Muller-Budack et~al.(2018)Muller-Budack, Pustu-Iren, and
  Ewerth}]{muller2018geolocation}
Eric Muller-Budack, Kader Pustu-Iren, and Ralph Ewerth. 2018.
\newblock Geolocation estimation of photos using a hierarchical model and scene
  classification.
\newblock In \emph{Proceedings of the European Conference on Computer Vision
  (ECCV)}, pages 563--579.

\bibitem[{Narasimhan et~al.(2018)Narasimhan, Barzilay, and
  Jaakkola}]{narasimhan2018grounding}
Karthik Narasimhan, Regina Barzilay, and Tommi Jaakkola. 2018.
\newblock Grounding language for transfer in deep reinforcement learning.
\newblock \emph{Journal of Artificial Intelligence Research}, 63:849--874.

\bibitem[{Neuhold et~al.(2017)Neuhold, Ollmann, Rota~Bul\`o, and
  Kontschieder}]{MVD2017}
Gerhard Neuhold, Tobias Ollmann, Samuel Rota~Bul\`o, and Peter Kontschieder.
  2017.
\newblock \href {https://www.mapillary.com/dataset/vistas} {The mapillary
  vistas dataset for semantic understanding of street scenes}.
\newblock In \emph{International Conference on Computer Vision (ICCV)}.

\bibitem[{Radford et~al.(2021)Radford, Kim, Hallacy, Ramesh, Goh, Agarwal,
  Sastry, Askell, Mishkin, Clark et~al.}]{radford2learning}
Alec Radford, Jong~Wook Kim, Chris Hallacy, Aditya Ramesh, Gabriel Goh,
  Sandhini Agarwal, Girish Sastry, Amanda Askell, Pamela Mishkin, Jack Clark,
  et~al. 2021.
\newblock Learning transferable visual models from natural language
  supervision.
\newblock \emph{arXiv:2103.00020}.

\bibitem[{Rupprecht et~al.(2018)Rupprecht, Laina, Navab, Hager, and
  Tombari}]{rupprecht2018guide}
Christian Rupprecht, Iro Laina, Nassir Navab, Gregory~D Hager, and Federico
  Tombari. 2018.
\newblock Guide me: Interacting with deep networks.
\newblock In \emph{Proceedings of the IEEE Conference on Computer Vision and
  Pattern Recognition}, pages 8551--8561.

\bibitem[{Theiner et~al.(2022)Theiner, M{\"u}ller-Budack, and
  Ewerth}]{theiner2022interpretable}
Jonas Theiner, Eric M{\"u}ller-Budack, and Ralph Ewerth. 2022.
\newblock Interpretable semantic photo geolocation.
\newblock In \emph{Proceedings of the IEEE/CVF Winter Conference on
  Applications of Computer Vision}, pages 750--760.

\bibitem[{Weyand et~al.(2016)Weyand, Kostrikov, and Philbin}]{weyand2016planet}
Tobias Weyand, Ilya Kostrikov, and James Philbin. 2016.
\newblock Planet-photo geolocation with convolutional neural networks.
\newblock In \emph{European Conference on Computer Vision}, pages 37--55.
  Springer.

\bibitem[{Yang et~al.(2021)Yang, Gan, Wang, Hu, Lu, Liu, and
  Wang}]{yang2021empirical}
Zhengyuan Yang, Zhe Gan, Jianfeng Wang, Xiaowei Hu, Yumao Lu, Zicheng Liu, and
  Lijuan Wang. 2021.
\newblock An empirical study of gpt-3 for few-shot knowledge-based vqa.
\newblock \emph{arXiv preprint arXiv:2109.05014}, 3(6):7.

\bibitem[{Zhong et~al.(2020)Zhong, Rockt{\"a}schel, and
  Grefenstette}]{Zhong2020-rtfm}
Victor Zhong, Tim Rockt{\"a}schel, and Edward Grefenstette. 2020.
\newblock {RTFM}: Generalising to novel environment dynamics via reading.
\newblock In \emph{International Conference on Learning Representations}.

\end{thebibliography}
